\documentclass[journal]{IEEEtran}

\usepackage{cite}
\usepackage{amsmath,amssymb,amsfonts}
\usepackage{algorithmic}
\usepackage{graphicx}
\usepackage{textcomp}
\usepackage{xcolor}
\usepackage{multirow}
\usepackage{adjustbox}
\usepackage{booktabs}
\usepackage{tikz}

\usepackage[hidelinks]{hyperref}
\usepackage{orcidlink}

\begin{document}

\title{Spatially Selective Self-Training for Unsupervised Building Change Detection}

\author{
Wafaa I. M. Hussin,
Zhi Lu,
Anas M. I. Mohammed,
Xiang Zhou,
Ratiba A. H. Abubaker,\\
and Zhenming Peng,~\IEEEmembership{Senior Member,~IEEE}%
\thanks{Wafaa I. M. Hussin, Xiang Zhou, and Zhenming Peng are with the School of Information and Communication Engineering, University of Electronic Science and Technology of China, Chengdu 611731, China (e-mails: wafaaibrahim20@gmail.com; zhouxiang@alu.uestc.edu.cn; zmpeng@uestc.edu.cn). Xiang Zhou is also with Chengdu Yaguang Electronic Co., Ltd.}%
\thanks{Zhi Lu is with the Laboratory of Intelligent Collaborative Computing, University of Electronic Science and Technology of China, Chengdu 611731, China (e-mail: zhilu@uestc.edu.cn).}%
\thanks{Anas M. I. Mohammed is with the School of Civil Engineering, University of Khartoum, Khartoum, Sudan (e-mail: anas66831@gmail.com).}%
\thanks{Ratiba A. H. Abubaker is with the National Energy Research Center, Ministry of Higher Education and Scientific Research, Khartoum, Sudan (e-mail: ratibanouralla@gmail.com).}%
}

\maketitle

\begin{abstract}
Unsupervised building change detection aims to learn building-change masks from unlabeled bi-temporal remote sensing images. Existing label-free methods often follow a discrepancy-to-mask paradigm, directly using temporal differences, frozen foundation-model responses, prompt-based outputs, or post-processing results as final change maps. Although these strategies provide annotation-free cues, they do not learn a task-specific building-change detector and remain vulnerable to the gap between generic temporal discrepancies and building-defined structural changes. In practice, such discrepancies are often noisy and task-irrelevant, as appearance shifts, registration errors, and non-building modifications can produce strong but misleading responses. To address this problem, we propose SST-CD, a spatially selective self-training framework that reformulates fully label-free building change detection as end-to-end detector learning under noisy pseudo supervision. SST-CD uses temporal discrepancies as candidate pseudo labels and trains the detector only on spatially reliable pixels, whose reliability is estimated by a local consistency criterion that filters inconsistent regions from supervision. To further stabilize noisy self-training, a lightweight feature adapter recalibrates bi-temporal features, while a prototype-based decoder produces compact change and no-change representations. Experiments on LEVIR-CD, WHU-CD, and DSIFN-CD show that SST-CD achieves F1 scores of 83.08\%, 91.69\%, and 86.60\%, respectively, outperforming existing unsupervised and label-free baselines.
\end{abstract}


\section{Introduction}
\label{sec:intro}
\IEEEPARstart{B}{uilding} change detection (BCD) aims to identify structural modifications, such as construction, demolition, and footprint expansion, from bi-temporal remote sensing images. It is fundamental for urban monitoring, map updating, and disaster assessment~\cite{ChangeDetectionTechniquesLu04}. With the development of deep neural networks, supervised BCD methods have achieved strong performance by learning discriminative bi-temporal representations from dense pixel-level annotations~\cite{li2024stade,11501963,S2LookingSatelliteSidelookingShen21,DSAnetNovelDeeplyDing21,HFANetHighFrequencyZheng22}.
However, dense change annotation is labor-intensive and requires domain expertise, especially when small buildings, dense urban layouts, and ambiguous boundaries make pixel-level labeling expensive and error-prone. This annotation bottleneck limits the scalability of supervised methods and motivates annotation-efficient change detection~\cite{SurveySampleEfficientDeepDing25}.

Existing annotation-efficient methods reduce reliance on dense masks in different ways. Semi-supervised methods use unlabeled image pairs with limited labeled samples, typically via pseudo-label learning, consistency regularization, or teacher-student training~\cite{DynamicallyUpdatedSemiSupervisedYuan24,AdaSemiCDAdaptiveSemiSupervisedRan25,IntegratingLocalGlobalZhang25}.
Weakly supervised methods replace dense masks with weaker task annotations and recover pixel-level predictions through localization or pseudo-mask generation~\cite{WeaklySupervisedBuildingMa24,TransWCDSceneAdaptiveJointZhao25,SWCDAccurateChangeTan25}.
Self-supervised methods learn transferable representations from unlabeled imagery, but often serve as pretraining and still require downstream adaptation or manual rules to produce final change maps~\cite{SSLChangeSelfSupervisedChangeZhao24,ICSFIntegratingInterModalZhang25,SelfsupervisedChangeDetectionChen22,SelfSupervisedPretrainingMultimodalityZhang23,S2CLearningNoiseresistantDing25}. Although these paradigms reduce annotation cost, most still depend on human-provided cues, supervised adaptation, or extra conversion rules to define building change.

Recent vision foundation models offer new opportunities for label-free change detection by providing generic structural and semantic priors~\cite{SegmentAnythingKirillov23,DINOv2LearningRobustOquab24,LearningTransferableVisualRadford21}. Existing label-free methods usually derive change maps from frozen features, promptable masks, or post-processed foundation-model responses~\cite{AnyChangeZheng24,SegmentChangeModelTan24,DynamicEarthHowFarLi26,UniVCDNewMethodZhu25}. Although annotation-free, these methods are largely training-free and prediction-oriented, making their outputs dependent on frozen responses and hand-designed conversion rules. In contrast, supervised adaptation methods improve task alignment by fine-tuning foundation models or adding task-specific decoders, but still require annotated data~\cite{MultiScaleRemoteSensingLiu26,FoundationModelDrivenSemanticShen26}. Therefore, fully label-free learning of a building-specific change detector remains underexplored.

This issue is central to fully unsupervised building change detection, where only unlabeled bi-temporal image pairs are available. Unlike single-image unsupervised segmentation~\cite{CutLERWang23,U2SegNiu23}, building change detection requires cross-temporal reasoning under a task-specific definition of structural change. Temporal discrepancies often fail to reflect true building changes, as nuisance variations may dominate while construction or demolition is sparse and spatially structured. Therefore, fully label-free building change detection demands a reliability-aware mechanism that converts noisy temporal discrepancies into effective supervision for task-specific detector learning.

This work revisits fully unsupervised building change detection as trainable detector learning rather than discrepancy-to-mask inference. Instead of treating temporal discrepancies as final predictions, we use them as candidate pseudo supervision for learning a task-specific building-change detector. The key challenge is therefore to identify which pseudo-labeled regions are reliable enough for optimization. Our insight is that reliable change and no-change regions are usually spatially coherent, whereas noisy responses tend to be fragmented or locally inconsistent. Based on this insight, we propose Selective Self-Training for Change Detection (SST-CD), a fully label-free framework for building-change detector learning.

SST-CD extracts structural features with a frozen foundation encoder, derives pseudo labels from bi-temporal feature discrepancies, and trains the detector only on locally reliable pixels selected by spatial consistency. To support this selective optimization, SST-CD introduces a lightweight feature adapter to recalibrate bi-temporal features before differencing and a prototype-based decoder to organize difference representations into compact change and no-change patterns. Together, these designs use foundation-model priors as pseudo supervision rather than fixed prediction outputs, enabling efficient detector learning from unlabeled image pairs.

The main contributions of this work are summarized as follows:
\begin{enumerate}
    \item We reformulate fully unsupervised building change detection from discrepancy-to-mask inference into trainable detector learning with spatially selective pseudo supervision.
    \item The detector incorporates a lightweight bi-temporal feature adapter to recalibrate features and a prototype-based decoder to encode compact change and no-change representations, supporting stable learning from noisy pseudo supervision.
    \item Extensive experiments on three public benchmarks demonstrate that SST-CD consistently outperforms existing unsupervised and label-free methods, demonstrating the effectiveness of selective pseudo-label optimization.
\end{enumerate}

\section{Related Work}
\label{sec:related_work}

\subsection{Annotation-Efficient and Unsupervised Change Detection}
Dense pixel-level annotation remains a major bottleneck for remote sensing change detection, motivating a growing body of annotation-efficient learning methods~\cite{SurveySampleEfficientDeepDing25}. Semi-supervised methods exploit unlabeled image pairs together with a small labeled subset, commonly through pseudo-label learning, consistency regularization, or teacher-student collaboration~\cite{SemiCDNetSemisupervisedConvolutionalPeng21, ReliableContrastiveLearningWang22, DynamicallyUpdatedSemiSupervisedYuan24, AdaSemiCDAdaptiveSemiSupervisedRan25, IntegratingLocalGlobalZhang25}. Weakly supervised methods replace dense masks with weaker task annotations and recover pixel-level predictions through weak-label localization, box-to-mask pseudo-label generation, or adversarial constraints~\cite{FullyConvolutionalChangeWu23,WeaklySupervisedBuildingMa24,ESAMCDFineTunedEfficientSAMWang24}. Self-supervised methods learn temporal or multi-modal representations from unlabeled imagery, but they are often used for representation pretraining or require additional decision rules to obtain final change maps~\cite{SelfsupervisedChangeDetectionChen22, SelfSupervisedPretrainingMultimodalityZhang23}. Although these paradigms reduce annotation cost, many of them still rely on human-provided task cues, downstream labeled adaptation, or manually specified conversion rules.

Fully unsupervised change detection is more aligned with the setting studied in this work, as it aims to infer change maps without manual change annotations. Classical UCD methods typically construct temporal discrepancy cues and separate changed from unchanged regions through thresholding, clustering, or statistical modeling~\cite{DigitalChangeDetectionSingh89, ChangeDetectionTechniquesLu04,TheoreticalFrameworkUnsupervisedBovolo07, UnsupervisedChangeDetectionCelik09}. Deep UCD methods improve this discrepancy space using learned representations, pretrained features, reconstruction objectives, cross-sensor mappings, or contrastive embeddings~\cite{UnsupervisedDeepChangeSaha19, ContentInvariantDualLearningFang22, UnsupervisedChangeDetectionNoh24, S2CLearningNoiseresistantDing25}.

Despite these advances, many existing methods still treat discrepancy separation as the terminal decision rule, where temporal cues are thresholded, clustered, or statistically partitioned into change maps. Such a formulation provides limited room for task-oriented learning from unlabeled data. Our method instead uses temporal discrepancies as provisional supervision, allowing a change detector to be trained rather than directly derived from hand-crafted or frozen discrepancy cues.

\subsection{Foundation Models for Change Detection}
Vision foundation models provide generic visual priors through promptable segmentation, language-aligned representation, and dense self-supervised features~\cite{SegmentAnythingKirillov23, LearningTransferableVisualRadford21, EmergingPropertiesSelfSupervisedCaron21,DINOv2LearningRobustOquab24}.
These priors have recently been introduced into remote sensing change detection under supervised or annotation-efficient settings. Existing methods typically use foundation models as auxiliary representation sources or semantic priors, and align them with the target CD task using pixel-level masks or weaker task labels~\cite{NewLearningParadigmLi24, AdaptingSegmentAnythingDing24, ChangeCLIPDong24, ESAMCDFineTunedEfficientSAMWang24}. While these studies demonstrate the usefulness of foundation-model representations, their task alignment still depends on explicit supervision.

Another line of work explores label-free or zero-shot CD by reusing foundation models as change-cue generators. Frozen masks, feature similarities, language-aligned scores, or promptable outputs are converted into change maps without manual annotations ~\cite{AnyChangeZheng24, SegmentChangeModelTan24, UniVCDNewMethodZhu25}. While appealing, these methods are mainly designed for open-vocabulary or promptable inference rather than unsupervised detector learning. Their predictions are therefore often obtained as post-hoc transformations of frozen foundation-model responses, making the results sensitive to the suitability of these responses for the target change category.

In contrast, our method uses foundation-model discrepancies to construct pseudo supervision rather than final predictions. By selecting locally reliable pseudo-labeled pixels for optimization, it learns a task-specific change detector from unlabeled bi-temporal pairs.

\subsection{Learning from Noisy Pseudo Supervision}
Pseudo-label learning is widely used in semi-supervised, weakly supervised, and label-free settings, but unreliable pseudo labels can reinforce erroneous predictions when treated as ground truth. Teacher-student consistency, confidence-regularized self-training, co-training, and reliable-sample selection have been developed to reduce error accumulation under noisy or model-generated supervision~\cite{MeanTeachersAreTarvainen17, ConfidenceRegularizedSelfTrainingZou19, CoteachingRobustTrainingHan18}. 
This problem becomes more pronounced in change detection, where strong visual differences may correspond to target changes or arise from illumination variation, seasonal appearance, shadows, and registration noise.

Recent annotation-efficient CD methods improve pseudo-label reliability through temporal stability modeling, dynamic screening, uncertainty-aware teacher updates, or cross-architecture collaboration~\cite{ReliableContrastiveLearningWang22, DynamicallyUpdatedSemiSupervisedYuan24, AdaSemiCDAdaptiveSemiSupervisedRan25, IntegratingLocalGlobalZhang25}. Most of these methods, however, are developed with some labeled samples or teacher signals that help anchor the semantic meaning of change. In fully label-free building CD, the pseudo labels themselves are derived from unlabeled bi-temporal discrepancies, making reliability estimation part of the learning problem.

Our method addresses this problem by estimating pseudo-label reliability from local spatial consistency. Rather than using all generated pseudo labels for training, it retains spatially coherent regions and excludes locally inconsistent ones from the loss. This makes pseudo-label selection the key mechanism for learning under noisy label-free supervision.

\section{Methodology}
\label{sec:methodology}
\subsection{Problem Formulation}
Given a bi-temporal remote sensing image pair $(I_a, I_b)$ acquired over the same area at two different times, where $I_a, I_b \in \mathbb{R}^{C \times H \times W}$, building change detection aims to predict a binary mask $Y \in \{0,1\}^{H \times W}$, where $Y_{ij}=1$ denotes a building-related change at pixel $(i,j)$ and $Y_{ij}=0$ otherwise. In the unsupervised setting, the training data consist of an unlabeled dataset $\mathcal{D}=\{(I_a^{(n)}, I_b^{(n)})\}_{n=1}^{N}$ without ground-truth change masks.

Unsupervised building change detection is inherently affected by a discrepancy between observable temporal differences and task-defined changes. Large image-level differences may result from acquisition-dependent appearance variations rather than from building construction or demolition. Consequently, pseudo supervision derived from raw temporal discrepancies provides an unreliable learning signal, since it may encode building-irrelevant variations instead of the structural changes of interest. 

Moreover, change is not solely determined by visual or semantic saliency, but by the target task. Some temporal transitions may be both visually prominent and semantically valid, while remaining irrelevant to building change detection. In this setting, only building-related structural transitions constitute positive changes. Without annotations, enforcing this task-specific distinction is difficult, and unsupervised models may be biased toward generic temporal changes rather than the desired building-specific changes.

\begin{figure*}[!t]
\centering
\includegraphics[width=0.95\textwidth]{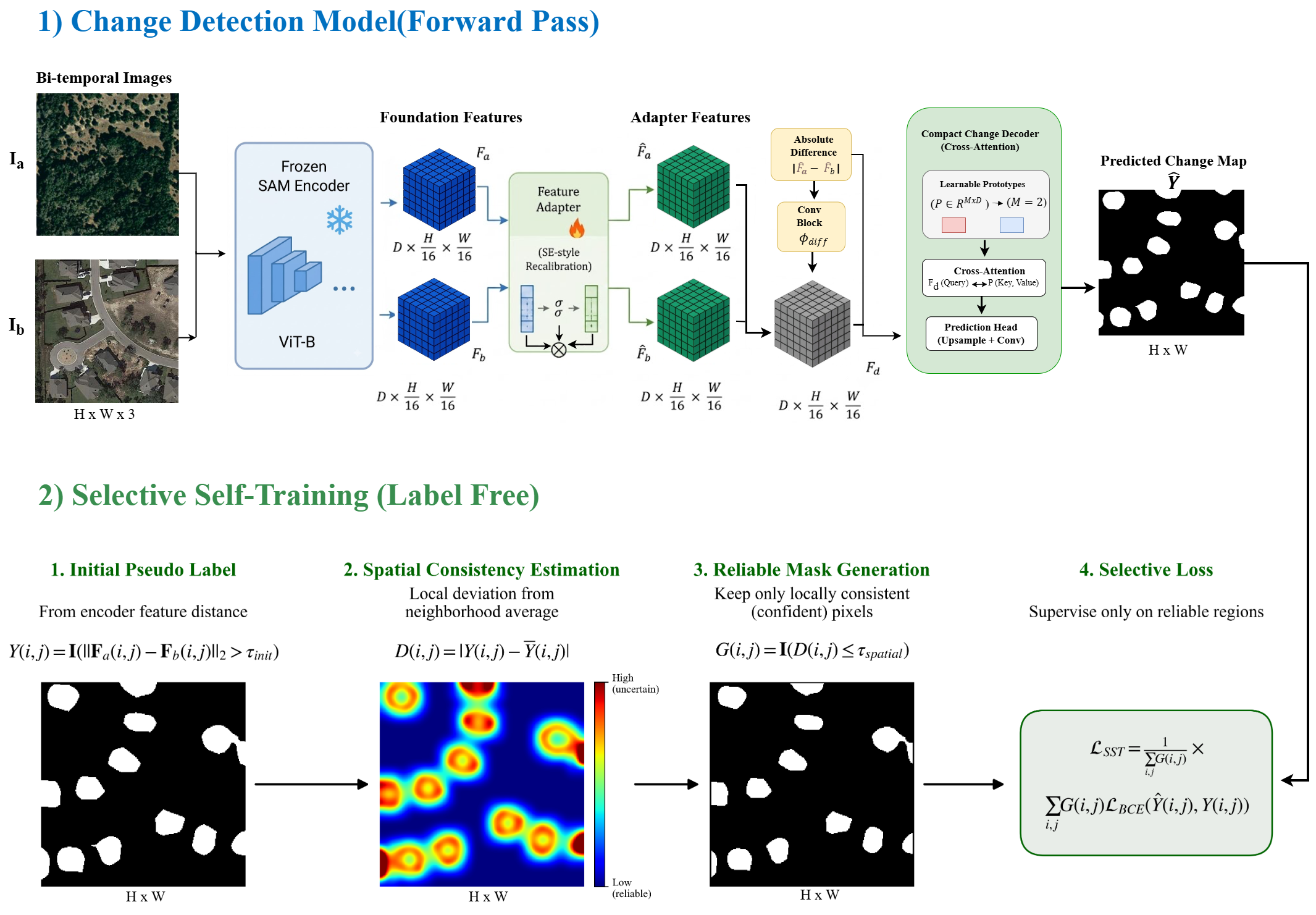} \caption{Framework of the proposed Selective Self-Training Change Detection (SST-CD) method. A frozen SAM encoder extracts foundation features from bi-temporal images ($I_a, I_b$). These are then calibrated by a Feature Adapter to suppress environmental noise. Their absolute difference is passed to a cross-attention decoder with $M=2$ learnable semantic prototypes (Change / No-Change) to predict the binary change mask $\hat{Y}$. To prevent confirmation bias, a Spatial Selective Loop generates a reliability mask ($G_s$) based on local spatial prediction variance with a fixed threshold $\tau$, restricting the self-training loss ($\mathcal{L}_{\mathrm{SST}}$) to only confident structural regions.}
\label{fig:architecture}
\end{figure*}

\subsection{Change Detection Model}
Our change detection model consists of a frozen image encoder, a lightweight feature adapter, and a Learnable Semantic Prototype decoder. Given a bi-temporal image pair, the model extracts structural features, adapts them to the remote sensing domain, and decodes a dense change mask.

\subsubsection{Frozen Image Encoder}
We adopt a Vision Transformer (ViT-B)~\cite{ImageWorth16x16Dosovitskiy21} as the image encoder $\mathcal{F}_{\text{encoder}}$, with weights initialized from SAM. Following prior work on large-scale vision pretraining, the encoder is designed to extract generic structural representations from high-resolution images. The vision encoder operates on a single image and produces a feature map that preserves rich structural and boundary information:
\begin{equation}\label{eq:feat}
\mathbf{F}_t = \mathcal{F}_{\text{encoder}}(I_t) \in \mathbb{R}^{D \times \frac{H}{16} \times \frac{W}{16}}, \, t \in \{a, b\}.
\end{equation}
Since the input resolution differs from the pretraining setting, we resize the positional embeddings using 2D interpolation~\cite{ImageWorth16x16Dosovitskiy21} to match the target patch-grid resolution.

\subsubsection{Feature Adapter}
Bi-temporal remote sensing images often contain appearance discrepancies caused by illumination, atmospheric conditions, or seasonal variations. These discrepancies can introduce global feature bias between $I_a$ and $I_b$, leading to spurious responses when the two feature maps are directly compared.

To reduce this bias, we introduce a lightweight feature adapter $\mathcal{F}_{\text{adapter}}$ for instance-level feature recalibration. The adapter is applied independently to each temporal feature map and uses global contextual statistics to modulate channel responses, inspired by squeeze-and-excitation (SE) modules~\cite{SqueezeandExcitationNetworksHu18}. For $t\in\{a,b\}$, we define
\begin{equation}
\mathbf{G}_t = \phi_{\mathrm{proj}}(\mathbf{F}_t),
\end{equation}
\begin{equation}
\hat{\mathbf{F}}_t
= \mathcal{F}_{\text{adapter}}(\mathbf{F}_t)
= \mathbf{G}_t \odot \sigma\big(\phi_{\mathrm{se}}(\mathbf{G}_t)\big),
\end{equation}
where $\phi_{\mathrm{proj}}(\cdot)$ denotes a lightweight convolutional projection, $\phi_{\mathrm{se}}(\cdot)$ denotes an SE-style recalibration function computed from globally pooled features, $\sigma(\cdot)$ is the sigmoid function, and $\odot$ denotes channel-wise multiplication.

\subsubsection{Prototype Decoder}
Given the adapted bi-temporal features $\hat{\mathbf{F}}_a$ and $\hat{\mathbf{F}}_b$, we first compute a difference representation:
\begin{equation}
\mathbf{F}_d = \phi_{\mathrm{diff}}  ( |\hat{\mathbf{F}}_a - \hat{\mathbf{F}}_b | )
\in \mathbb{R}^{D \times \frac{H}{16} \times \frac{W}{16}},
\end{equation}
where $\phi_{\mathrm{diff}}(\cdot)$ denotes a convolutional block.

Instead of directly predicting the change mask from $\mathbf{F}_d$, we introduce a Learnable Semantic Prototype (LSP) decoder~\cite{snell2017prototypical} to regularize the difference representation. Specifically, we maintain a set of learnable prototypes $\mathbf{P}\in\mathbb{R}^{M\times D}$ as anchors in the difference-feature space. The spatial dimensions of $\mathbf{F}_d$ are flattened into a sequence of tokens and used as queries, while the prototypes serve as keys and values:

\begin{equation}
\mathbf{F}_{\mathrm{proto}}
=
\mathrm{CrossAttn}(\mathbf{F}_d,\mathbf{P},\mathbf{P}),
\end{equation}
where $\mathrm{CrossAttn}(\cdot)$ denotes a standard cross-attention operation, with implicit flattening and reshaping of the spatial feature map.

Finally, $\mathbf{F}_{\mathrm{proto}}$ is fed into a prediction head $\mathcal{H}_{\mathrm{pred}}(\cdot)$ for upsampling and mask prediction:
\begin{equation}
\hat{Y}
=
\mathcal{H}_{\mathrm{pred}}(\mathbf{F}_{\mathrm{proto}})
\in \mathbb{R}^{H\times W}.
\end{equation}

This design encourages the decoder to represent difference features through a small set of learnable prototypes, thereby promoting compact and discriminative representations. The prototypes are optimized end-to-end and provide an adaptive basis for modeling recurring patterns in the difference-feature space. In our implementation, we set $M=2$ to represent the two target states, i.e., change and no-change.

\subsection{Selective Self-Training}
\label{sec:self_training}
Unsupervised building change detection typically relies on pseudo-supervision derived from unlabeled bi-temporal image pairs. Let $Y \in \{0,1\}^{H \times W}$ denote the static pseudo-label map generated from foundation model priors. Since such pseudo-labels are inevitably noisy, indiscriminate supervision over all pixels may propagate unreliable signals and bias the model toward pseudo-label artifacts.

To address this issue, we propose a selective training mechanism that filters unreliable pixels according to the spatial uncertainty of $Y$. Unlike prediction-confidence-based selection, our selection mask is computed only from the pseudo-label map itself and is independent of the model's current output. This design reduces the risk of confirmation bias by preventing the model from using its own potentially erroneous predictions to determine which pixels should supervise training.

\subsubsection{Pseudo-Label Generation}
We first construct a static pseudo-label map $Y \in \{0,1\}^{H \times W}$ from the frozen SAM encoder. Given a bi-temporal image pair, the encoder extracts feature maps $\mathbf{F}_a$ and $\mathbf{F}_b$ for the two temporal images. We then compute the pixel-wise feature discrepancy and obtain an initial binary change mask by applying a fixed threshold $\tau_{\mathrm{init}}$:
\begin{equation}
Y(i,j) = \mathbb{I}\bigl( \|\mathbf{F}_a(i,j) - \mathbf{F}_b(i,j)\|_2 > \tau_{\mathrm{init}} \bigr).
\end{equation}
This initial mask serves as noisy pseudo-supervision. It captures generic structural discrepancies between the two images, but may also include task-irrelevant variations such as illumination changes, shadows, vegetation differences, and registration noise. 

\subsubsection{Spatial Consistency Mask}
We evaluate the reliability of each pseudo-labeled pixel according to its local spatial consistency. In the pseudo-label map, genuine building changes usually appear as locally coherent regions, where pixels inside changed building footprints share consistent labels. Stable unchanged areas also exhibit similar local consistency. In contrast, spurious responses caused by shadows, vegetation variations, registration errors, or feature-difference noise are more likely to appear around boundaries or as isolated fragments. Therefore, local spatial consistency provides a simple and effective cue for selecting reliable pseudo-label supervision.

To quantify local consistency, we compute the average pseudo-label response within a neighborhood centered at each pixel:
\begin{equation}
\overline{Y}(i,j)=
\frac{1}{|\mathcal{N}_{ij}|}
\sum_{(m,n)\in\mathcal{N}_{ij}} Y(m,n),
\end{equation}
where $\mathcal{N}_{ij}$ denotes a local neighborhood centered at pixel $(i,j)$. In our implementation, $\mathcal{N}_{ij}$ is set to a $5\times5$ window.

We then define the local inconsistency score as the deviation between the pseudo label and its neighborhood average:
\begin{equation}
D(i,j)=\left|Y(i,j)-\overline{Y}(i,j)\right|.
\end{equation}
A smaller value of $D(i,j)$ indicates stronger agreement between a pixel and its surrounding pseudo labels, suggesting higher reliability for supervision. Conversely, a larger value indicates local label inconsistency, which often corresponds to ambiguous building boundaries, isolated noisy responses, or fragmented non-building structures. Based on this score, we construct a spatial consistency mask to retain locally coherent pixels and suppress spatially ambiguous ones during training.

\subsubsection{Selective Loss Masking}  
Based on the local inconsistency score, we define a binary selection mask $G$ to retain pixels with sufficiently high spatial consistency:
\begin{equation}\label{eq:consistancy}
G(i,j) = \mathbb{I}( D(i,j) \le \tau_{\mathrm{spatial}}),
\end{equation}
where $\tau_{\mathrm{spatial}}$ is a consistency threshold. 
Pixels with small inconsistency scores are treated as reliable training pixels, whereas pixels with large scores are excluded from the loss.

The selective self-training loss is then computed only over the selected pixels:
\begin{equation}
\mathcal{L}_{\mathrm{SST}} = \frac{1}{\sum_{i,j} G(i,j)}\sum_{i,j} G(i,j)\ell_{\mathrm{BCE}}(\hat{Y}(i,j), Y(i,j)),
\end{equation}
where $\ell_{\mathrm{BCE}}(p,y)=-[y\log p + (1-y)\log(1-p)]$ and $\hat{Y}$ denotes the model's prediction. The selection mask is computed from the static pseudo-label map and remains fixed during training. It restricts the loss to locally consistent pseudo-label regions, thereby reducing the influence of ambiguous boundaries and isolated noisy responses on model optimization.

\subsection{Test-Time Augmentation}
Remote sensing images are captured from a nadir viewpoint without a canonical orientation. As a result, building changes are expected to be equivariant to rotations and flips. To reduce orientation-specific biases at inference, we apply test-time geometric augmentation and aggregate predictions over multiple transformed views.

Given a bi-temporal pair $(I_a, I_b)$, we use a set of $N=8$ planar transformations $\{\mathcal{T}_k\}_{k=1}^{N}$, comprising four $90^\circ$ rotations and their horizontal flips. Let $f_\theta$ denote the entire change detection network parameterized by $\theta$. For each transform, we run the network on the transformed inputs and then invert the transform to align the output to the original coordinate frame:
\begin{equation}
\tilde{\mathbf{P}}_k = \mathcal{T}_k^{-1}(f_\theta(\mathcal{T}_k(I_a, I_b))).
\end{equation}
The final prediction is obtained by averaging the transformed outputs, followed by a lightweight morphological post-processing step to suppress isolated noise:
\begin{equation}
\mathbf{P}_{\mathrm{final}} = \mathrm{Morph} \left( \frac{1}{N} \sum_{k=1}^{N} \tilde{\mathbf{P}}_k \right),
\end{equation}
where $\mathrm{Morph}(\cdot)$ denotes a standard opening-closing morphological operation that removes isolated pixels and enforces local spatial coherence in the predicted change map. The operation is applied only at inference time and does not affect model training.

\section{Experiments}\label{sec:EXPERIMENTS}

\subsection{Datasets}
We evaluate our method on three public building change detection datasets: LEVIR-CD~\cite{SpatialTemporalAttentionBasedMethodChen20}, WHU-CD~\cite{FullyConvolutionalNetworksJi19}, and DSIFN-CD~\cite{DeeplySupervisedImageZhang20}.

\textbf{LEVIR-CD} contains 637 pairs of bi-temporal aerial images collected from Google Earth. Each image has a spatial resolution of 0.5,m and a size of $1024 \times 1024$ pixels. Following common practice~\cite{RemoteSensingImageChen22}, the images are cropped into non-overlapping $256 \times 256$ patches, resulting in 7,120 training pairs, 1,024 validation pairs, and 2,048 test pairs.

\textbf{WHU-CD} is derived from very high-resolution aerial images over Christchurch, New Zealand, capturing building changes between two acquisition dates. The original images have a spatial resolution of 7.5,cm and a size of $32{,}507 \times 15{,}354$ pixels. Following previous work~\cite{RemoteSensingImageChen22}, we crop the images into non-overlapping $256 \times 256$ patches, resulting in 6,096 training pairs, 762 validation pairs, and 762 test pairs.

\textbf{DSIFN-CD} consists of high-resolution bi-temporal satellite images collected from six major cities in China, covering heterogeneous land-cover changes including buildings, vegetation, roads, and water bodies. Using the official split and cropping each image into four $256 \times 256$ patches, we obtain 14,400 training pairs, 1,360 validation pairs, and 192 test pairs.

\begin{table*}[!t]
\centering
\caption{Comparison results for building change detection.}
\label{tab:comparison_final}
\setlength{\tabcolsep}{7.5pt}
\renewcommand{\arraystretch}{1.08}
\begin{tabular}{lccccccccc}
\hline
\multirow{2}{*}{\textbf{Method}} &
\multicolumn{3}{c}{\textbf{LEVIR-CD}} &
\multicolumn{3}{c}{\textbf{WHU-CD}} &
\multicolumn{3}{c}{\textbf{DSIFN-CD}} \\
\cline{2-4} \cline{5-7} \cline{8-10}
& F1 & IoU & mIoU & F1 & IoU & mIoU & F1 & IoU & mIoU \\
\hline
BIT~\cite{RemoteSensingImageChen22}
& 89.81 & 81.50 & 90.18 & 90.69 & 82.97 & 91.11 & 87.91 & 78.43 & 86.92 \\
\hline
PCA-KM~\cite{UnsupervisedChangeDetectionCelik09}&  9.29 &  4.87 & 33.78 &  5.67 &  2.92 & 34.90 & 42.49 & 26.98 & 38.49 \\
CVA~\cite{id21}&  8.10 &  4.22 & 33.82 & 16.42 &  8.94 & 43.00 & 38.23 & 23.63 & 37.34 \\
DCVA~\cite{UnsupervisedChangeDetectionCelik09}&  12.43 &  6.63 & 42.63 & 16.49 &  8.99 & 48.04 & 51.61 & 34.78 & 23.95  \\
CDRL \cite{id19}  & 11.53 & 6.12 & 45.08 & 30.16 & 17.76 & 56.26 & - & - &- \\
UniVCD~\cite{UniVCDNewMethodZhu25}  & 79.90 & 66.53 & 82.20 & 83.10 & 71.09 & 84.91 & 79.77 & 66.35 & 79.32 \\
\hline
SAM~\cite{SegmentAnythingKirillov23} & 20.06 & 11.15 & 49.92 & 25.08 & 14.34 & 51.25 & 38.72 & 24.01 & 36.46\\
DINOv2~\cite{DINOv2LearningRobustOquab24} & 16.89 &  9.22 & 33.51 & 15.85 &  8.61 & 33.19 & 54.26 & 37.23 & 42.70 \\
\midrule 
SST-CD  & \textbf{83.08} & \textbf{71.06} & \textbf{84.64}
                & \textbf{91.69} & \textbf{84.65} & \textbf{92.00}
                & \textbf{86.60} & \textbf{76.37} & \textbf{85.53} \\
\bottomrule
\end{tabular}
\end{table*}

\subsection{Baselines}
We compare SST-CD with representative building change detection methods spanning different supervision levels and model priors, including conventional unsupervised methods, deep unsupervised methods, foundation-model-based label-free methods, and a fully supervised reference. Specifically, PCA-KM~\cite{UnsupervisedChangeDetectionCelik09} and CVA~\cite{id21} are included as classical unsupervised baselines based on hand-crafted difference representations and clustering or thresholding. DCVA~\cite{UnsupervisedChangeDetectionCelik09} and CDRL~\cite{id19} are further considered as deep unsupervised baselines that exploit learned feature representations.

For foundation-model-based comparison, we include UniVCD~\cite{UniVCDNewMethodZhu25}, a representative open-vocabulary change detection method. We also construct two direct zero-shot baselines using SAM~\cite{SegmentAnythingKirillov23} and DINOv2~\cite{DINOv2LearningRobustOquab24}. For each zero-shot baseline, frozen foundation-model features from the bi-temporal images are differenced pixel-wise and thresholded by Otsu's method to obtain the final binary change mask. In SST-CD, these masks are instead used as noisy initial pseudo-labels, from which reliable regions are selected for task-specific detector learning.

Finally, BIT~\cite{RemoteSensingImageChen22} is reported as a fully supervised reference to indicate the remaining gap between annotation-free learning and supervised change detection.

\subsection{Evaluation Metrics}
We evaluate binary mask segmentation using pixel-level metrics. Let $\mathrm{TP}$, $\mathrm{FP}$, $\mathrm{FN}$, and $\mathrm{TN}$ denote the numbers of true positives, false positives, false negatives, and true negatives for the foreground (change) class. Precision (P), recall (R), F1 score, and Intersection over Union (IoU) are defined as:
\begin{equation}
\mathrm{P} = \frac{\mathrm{TP}}{\mathrm{TP} + \mathrm{FP}}, \,\,
\mathrm{R} = \frac{\mathrm{TP}}{\mathrm{TP} + \mathrm{FN}}, \,\,
\mathrm{F1} = \frac{2 \mathrm{P}\mathrm{R}}{\mathrm{P} + \mathrm{R}}, \,\,
\end{equation}

\begin{equation}
\mathrm{IoU} = \frac{\mathrm{TP}}{\mathrm{TP} + \mathrm{FP} + \mathrm{FN}}.
\end{equation}
We additionally report mean IoU (mIoU), computed as the average IoU over foreground and background classes:
\begin{equation}
\mathrm{mIoU} =
\frac{1}{2}
 (
\frac{\mathrm{TP}}{\mathrm{TP} + \mathrm{FP} + \mathrm{FN}} +
\frac{\mathrm{TN}}{\mathrm{TN} + \mathrm{FP} + \mathrm{FN}}
 ).
\end{equation}
All metrics are reported in percentage.

\subsection{Implementation Details}
Our model is implemented using the PyTorch~\cite{id16} and trained on a single NVIDIA TITAN RTX 3090 GPU (24 GB). The foundation feature extractor of our framework utilizes the Segment Anything Model (SAM) with a ViT-B backbone, which is initialized with pre-trained weights. The input bi-temporal images are processed at a spatial resolution of $256 \times 256$ pixels. During training, we employ standard data augmentation techniques, specifically random cropping alongside horizontal and vertical flipping. The network is optimized using the AdamW~\cite{id17} optimizer with an initial learning rate of $1\times10^{-5}$ and a weight decay of $1\times10^{-2}$. The training process is conducted for 100 epochs with a batch size of 4. During the inference phase, we apply an 8-view Test-Time Augmentation (TTA) strategy encompassing multiple rotations and flips to enhance prediction robustness.

\begin{figure*}[!t]
\centering
\includegraphics[width=0.98\linewidth]{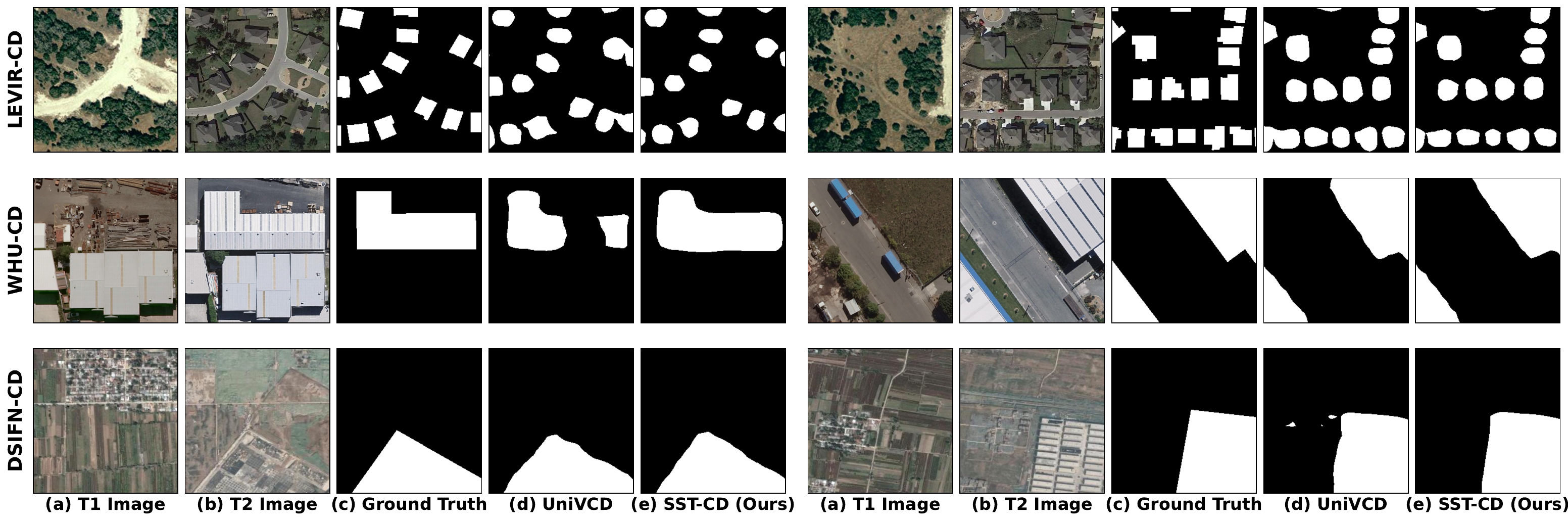}
\caption{Qualitative comparisons of the LEVIR-CD, WHU-CD, and DSIFN-CD datasets. The columns represent: (a) T1 Image, (b) T2 Image, (c) Ground Truth, (d) UniVCD, (e) SST-CD(Ours)}
\label{fig.performance_comparison}
\end{figure*}

\subsection{Overall Performance}
\label{sec:overall_performance}
Table~\ref{tab:comparison_final} reports the quantitative comparison on LEVIR-CD, WHU-CD, and DSIFN-CD. Among the unsupervised and label-free methods, SST-CD achieves the best overall performance on all three benchmarks. It obtains F1 scores of 83.08\%, 91.69\%, and 86.60\% on LEVIR-CD, WHU-CD, and DSIFN-CD, respectively. Compared with UniVCD, the strongest label-free baseline in Table~\ref{tab:comparison_final}, SST-CD improves the F1 score by 3.18\%, 8.59\%, and 6.83\% on the three datasets. These results indicate that the proposed framework can effectively learn building-change representations from unlabeled bi-temporal image pairs

We also evaluate a training-free feature-differencing baseline using frozen foundation-model features. For SAM and DINOv2, spatial features are extracted from the two temporal images, and their pixel-wise absolute differences are thresholded by Otsu's method to obtain binary change maps. This direct strategy performs poorly on LEVIR-CD, with F1 scores of 20.06\% for SAM and 16.89\% for DINOv2. Similar limitations are observed on WHU-CD and DSIFN-CD. These results indicate that raw feature discrepancies are insufficient for reliable change prediction. Although frozen foundation features provide useful structural cues, their temporal differences may also capture appearance variations, shadows, vegetation, and registration noise.

Rather than taking frozen feature discrepancies as final predictions, SST-CD uses them as candidate pseudo supervision and trains the detector only with reliable pseudo-labeled pixels. This selective optimization strategy enables more effective use of label-free change cues and explains the consistent improvements over existing unsupervised and label-free baselines.

Qualitative results in Fig.~\ref{fig.performance_comparison} further support these findings. Across the three datasets, SST-CD produces change masks that better align with the ground truth, reduces false connections between adjacent buildings, and yields more complete and coherent changed regions. In contrast, UniVCD often produces fragmented or irregular masks. These observations are consistent with the quantitative improvements, confirming that SST-CD enhances both pixel-level accuracy and structural consistency.

\begin{figure}[!t]
\centering
\includegraphics[width=1\linewidth]{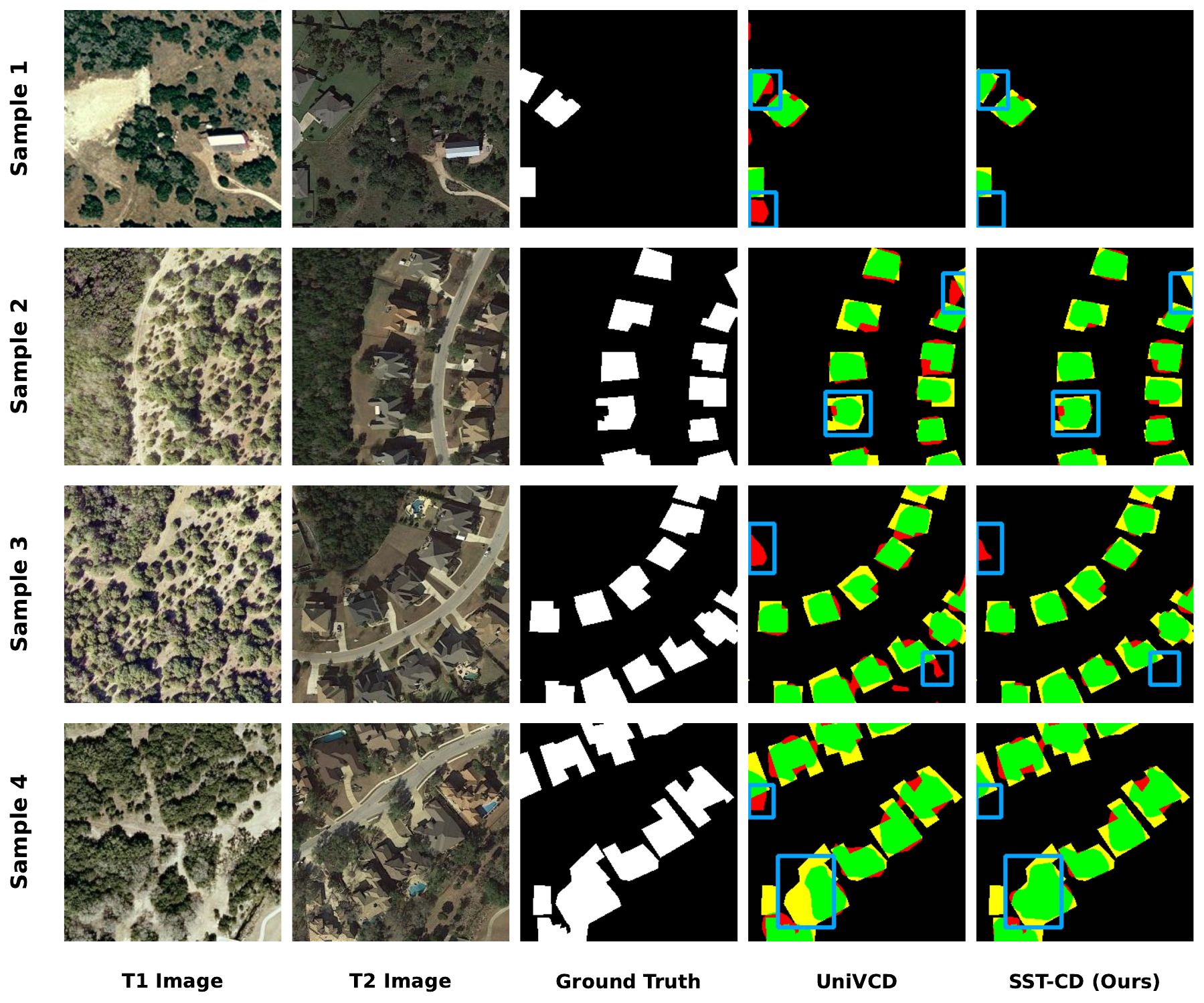}
\caption{Qualitative comparison on representative LEVIR-CD samples. From left to right, each column shows the pre-change image, post-change image, ground truth, and error maps of UniVCD and SST-CD. In the error maps, green, red, and yellow denote true positives, false positives, and false negatives, respectively. Blue boxes highlight regions where SST-CD better separates nearby small buildings and reduces missed detections.}
\label{fig.visual_comparison_two_rows}
\end{figure}

Fig.~\ref{fig.visual_comparison_two_rows} presents multiple representative LEVIR-CD examples using error-map visualization. The columns show the pre-change image, post-change image, ground-truth mask, and the error maps of UniVCD and SST-CD. UniVCD produces several false-positive regions between adjacent buildings, which leads to undesired connections between nearby changed instances. In contrast, SST-CD noticeably reduces these inter-building false positives and better preserves the separation of small changed regions. Moreover, the remaining errors of SST-CD are mainly concentrated around object boundaries rather than large non-building regions, suggesting improved spatial localization and structural coherence in dense building-change prediction.

\begin{table}[!t]
\centering
\caption{Ablation results on LEVIR-CD and WHU-CD.}
\label{tab:ablation_study}
\footnotesize
\setlength{\tabcolsep}{3.5pt}
\begin{tabular}{lccc|ccc}
\hline
\multirow{2}{*}{Method} & \multicolumn{3}{c|}{LEVIR-CD} & \multicolumn{3}{c}{WHU-CD} \\
\cline{2-4} \cline{5-7}
& F1 & IoU & mIoU & F1 & IoU & mIoU \\
\hline
Base model 
& 75.76 & 60.97 & 79.27 
& 81.36 & 68.56 & 83.61 \\
+ Feature Adapter 
& 78.73 & 64.92 & 81.38 
& 88.89 & 80.00 & 89.56 \\
+ Prototype Decoder 
& 79.85 & 66.47 & 82.16 
& 88.47 & 79.33 & 89.21 \\
+ Selective Loss 
& 81.55 & 68.84 & 83.45 
& 90.39 & 82.46 & 90.85 \\
+ Test Time Augmentation
& \textbf{83.08} & \textbf{71.06} & \textbf{84.64} 
& \textbf{91.69} & \textbf{84.65} & \textbf{92.00} \\
\hline
\end{tabular}
\end{table}

\subsection{Ablation Study}
Table~\ref{tab:ablation_study} reports the component ablation results on LEVIR-CD and WHU-CD. The base model consists of the frozen encoder, a simple interaction decoder, and the standard BCE loss trained with the initial pseudo labels. Compared with directly thresholding frozen features in Table~\ref{tab:comparison_final}, the base model already achieves much higher performance, indicating that learning a detector from pseudo supervision is more effective than using raw feature discrepancies as final predictions.

Adding the feature adapter consistently improves performance on both datasets. The IoU increases from 60.97\% to 64.92\% on LEVIR-CD and from 68.56\% to 80.00\% on WHU-CD. This indicates that recalibrating bi-temporal features before differencing helps reduce appearance-induced discrepancies and produces more discriminative change representations. The prototype decoder brings a modest gain on LEVIR-CD and remains comparable to the feature adapter variant on WHU-CD. This suggests that prototype-based decoding is helpful, but stable improvement still depends on reliable pseudo-label selection. After introducing the selective loss, the performance improves consistently on both datasets. the IoU increases to 68.84\% on LEVIR-CD and 82.46\% on WHU-CD. This indicates that excluding locally unreliable pseudo-labeled pixels helps the model learn from cleaner supervision. Finally, test-time augmentation further improves the results, reaching 83.08\% F1 on LEVIR-CD and 91.69\% F1 on WHU-CD. Since test-time augmentation is applied only during inference, this gain reflects improved prediction stability under geometric transformations.

\subsection{Method Analysis}
\subsubsection{Feature Discriminability Analysis}
To further analyze the effect of the feature adapter, we measure the separation between changed and unchanged regions in the feature-difference space. Let $\bar{d}_{c}$ and $\bar{d}_{u}$ denote the average $\ell_1$ distance between $\mathbf{F}a$ and $\mathbf{F}b$ over changed and unchanged pixels, respectively. The discriminability ratio is defined as
\begin{equation}
    R_d = {\bar{d}_{c}}/{\bar{d}_{u}} .
\end{equation}
A larger $R_d$ indicates that changed regions have stronger feature responses relative to unchanged regions.

\begin{table}[!t]
\centering
\caption{Feature discriminability analysis. A larger $R_d$ indicates better separation between changed and unchanged regions.}
\label{tab:fab_discriminability}
\footnotesize
\begin{tabular}{lccc}
\hline
Dataset & w/o adapter & w/ adapter & Gain \\
\hline
WHU-CD   & 1.38 & \textbf{1.81} & +31.3\% \\
LEVIR-CD & 1.32 & \textbf{1.49} & +12.6\% \\
\hline
\end{tabular}
\end{table}

\begin{table}[!t]
\centering
\caption{Comparison of pixel selection strategies on WHU-CD under the same retention ratio of 99.37\%.}
\label{tab:random_vs_selective}
\footnotesize
\setlength{\tabcolsep}{8pt}
\begin{tabular}{lc}
\hline
Strategy & F1 (\%) \\
\hline
Random masking & 91.63 \\
Selective masking & \textbf{91.98} \\
\hline
\end{tabular}
\end{table}
Table~\ref{tab:fab_discriminability} shows that the feature adapter increases $R_d$ on both datasets, from 1.38 to 1.81 on WHU-CD and from 1.32 to 1.49 on LEVIR-CD. This suggests that the adapted features provide better separation between changed and unchanged regions, which is consistent with the improvement observed after adding the feature adapter in Table~\ref{tab:ablation_study}.

\begin{figure}[!t]
\centering
\includegraphics[width=1\linewidth]{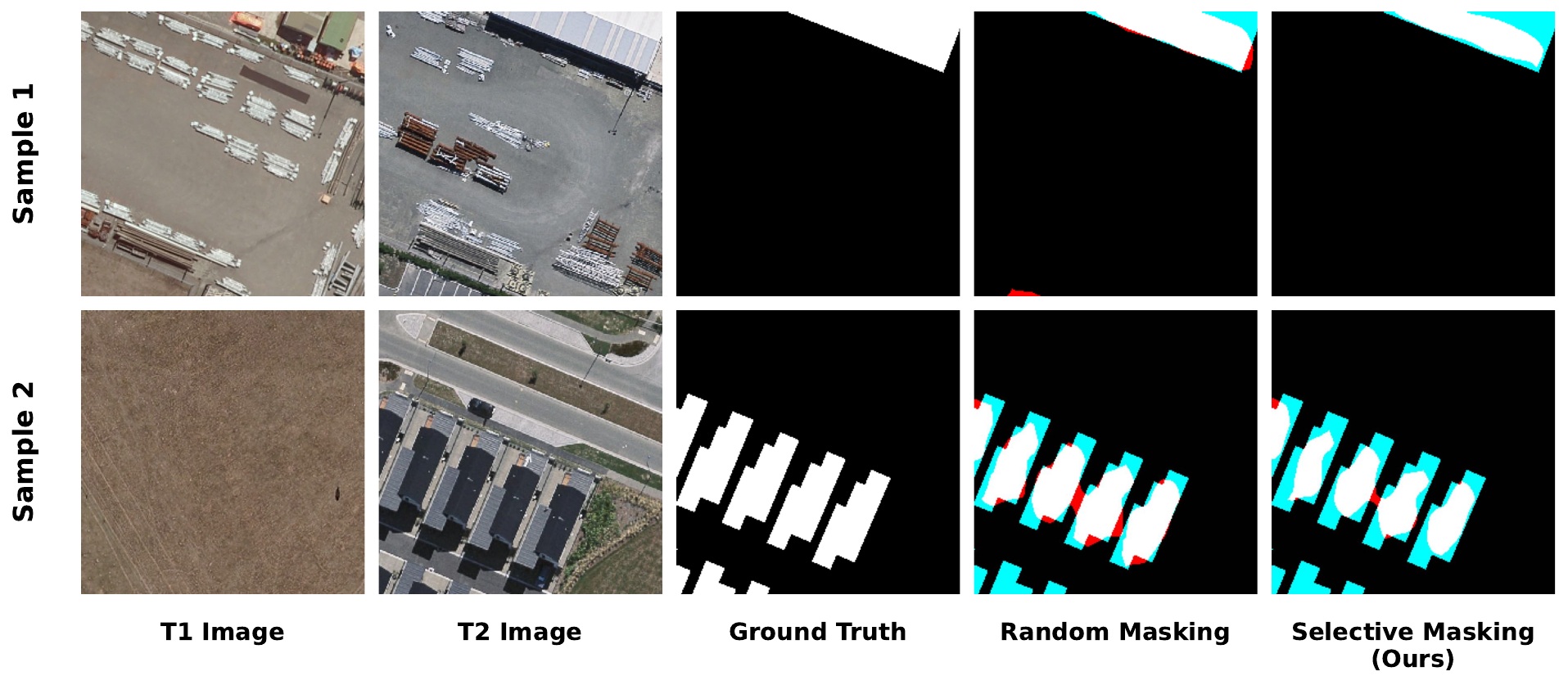}
\caption{Error-map comparison between random masking and selective masking on WHU-CD. Selective masking reduces local boundary errors and isolated false positives while preserving more coherent changed regions.}
\label{fig:visual_ablation_random_selective}
\end{figure}

\subsubsection{Selective Loss versus Random Masking}
To examine whether the benefit of selective loss comes from spatially guided pixel selection rather than pixel sparsity alone, we compare it with random masking under the same retention ratio.

Table~\ref{tab:random_vs_selective} shows that selective masking achieves a slightly higher F1 score than random masking, improving from 91.63\% to 91.98\%. This indicates that the gain is not merely due to reducing the number of supervised pixels. The error maps in Fig.~\ref{fig:visual_ablation_random_selective} further show that random masking tends to introduce local boundary errors and scattered false positives, whereas selective masking produces more coherent changed regions with fewer structural errors. These results suggest that spatially guided pixel selection is more effective than random removal for preserving geometric consistency.

\subsubsection{Sensitivity to Consistency Threshold}
We further analyze the sensitivity of the consistency threshold $\tau_{\mathrm{spatial}}$ in Eq.~\ref{eq:consistancy}. As shown in Table~\ref{tab:combined_tau_sensitivity}, the performance remains relatively stable across different thresholds, indicating that the selective loss is not highly sensitive to a single threshold value.

On LEVIR-CD, the best performance is obtained at $\tau_{\mathrm{spatial}}=0.25$, while WHU-CD achieves the highest F1 and IoU at $\tau_{\mathrm{spatial}}=0.80$. This difference suggests that the appropriate filtering strength may vary across datasets. Smaller and denser building regions tend to benefit from stricter selection, whereas larger and more continuous regions can tolerate a more relaxed threshold. In both datasets, setting $\tau_{\mathrm{spatial}}=1.00$ is not optimal, indicating that spatial consistency-based selection remains useful when training with noisy pseudo labels.

\begin{table}[!t]
\centering
\caption{Sensitivity to the consistency threshold $\tau_{\mathrm{spatial}}$.}
\label{tab:combined_tau_sensitivity}
\footnotesize
\begin{tabular}{ccc|cc}
\hline
\multirow{2}{*}{$\tau_{\mathrm{spatial}}$}
& \multicolumn{2}{c}{LEVIR-CD}
& \multicolumn{2}{c}{WHU-CD} \\
\cline{2-3} \cline{4-5}
& F1 & IoU & F1 & IoU \\
\hline
0.05 & 82.02 & 69.52 & 91.45 & 84.24 \\
0.20 & 82.12 & 69.67 & 91.68 & 84.64 \\
0.25 & \textbf{82.85} & \textbf{70.72} & 91.39 & 84.14 \\
0.40 & 82.66 & 70.44 & 91.90 & 85.02 \\
0.60 & 82.81 & 70.66 & 91.94 & 85.09 \\
0.80 & 82.35 & 69.99 & \textbf{91.98} & \textbf{85.16} \\
1.00 & 81.67 & 69.02 & 91.86 & 84.95 \\
\hline
\end{tabular}
\end{table}

\subsubsection{Prediction Dynamics under Selective Supervision}
\begin{figure}[!t]
\centering
\includegraphics[width=0.90\linewidth]{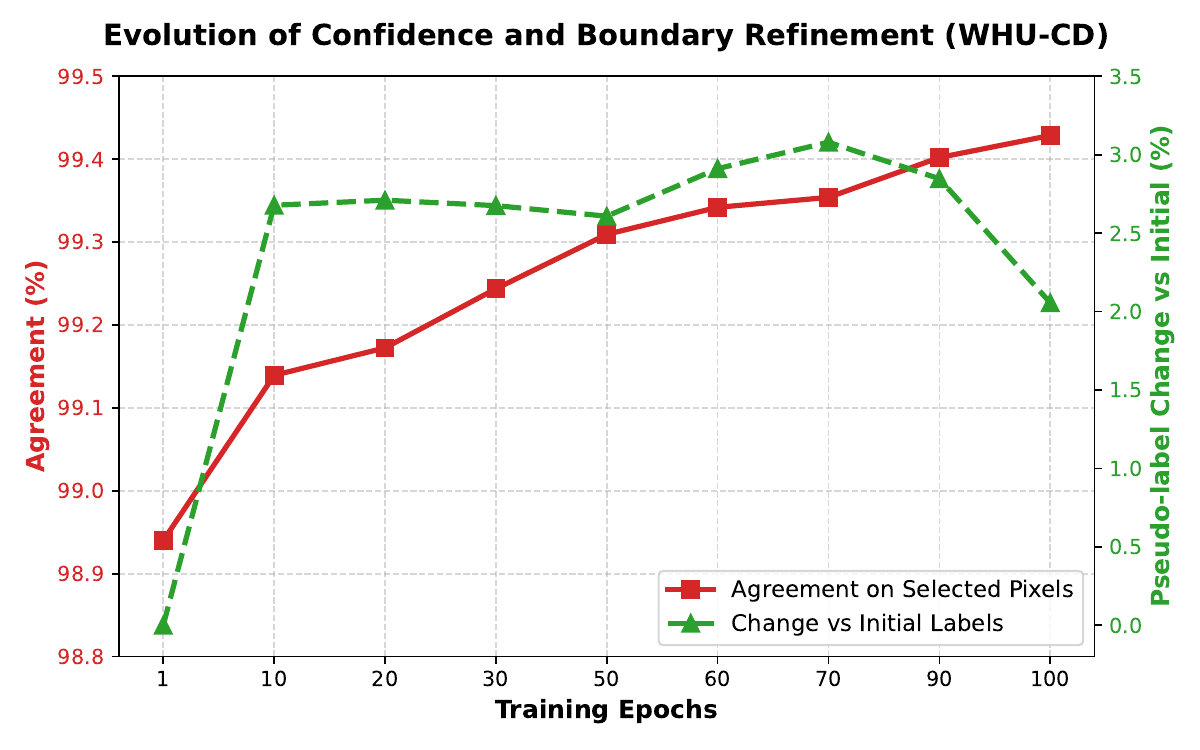}
\caption{Prediction dynamics under selective supervision on WHU-CD. The red curve denotes the agreement between model predictions and selected pseudo labels, while the green curve denotes the percentage of predictions changed from the initial pseudo labels.}
\label{fig:evolution_confidence}
\end{figure}

Fig.~\ref{fig:evolution_confidence} shows the prediction dynamics of SST-CD on WHU-CD over 100 training epochs. The red curve denotes the agreement rate between the binarized model predictions and the selected pseudo labels. It increases from 98.94\% to 99.43\%, indicating stable convergence under selective supervision. The green curve denotes the percentage of pixels whose predictions differ from the initial pseudo labels. This value remains limited, mostly around 2\%--3\%, suggesting that training introduces only moderate changes to the initial pseudo supervision.

\section{Conclusion}
\label{sec:conclusion}
In this paper, we proposed SST-CD, a selective self-training framework for unsupervised building change detection. Instead of directly converting temporal discrepancies into final change maps, SST-CD treats them as noisy candidate supervision and trains a change detector using only spatially reliable pseudo-labeled pixels. This formulation shifts label-free change detection from discrepancy-based inference to selective detector learning under noisy supervision. Experiments on LEVIR-CD, WHU-CD, and DSIFN-CD show that SST-CD achieves F1 scores of 83.08\%, 91.69\%, and 86.60\%, respectively, outperforming existing unsupervised and label-free baselines. These results demonstrate that controlling the supervision signal during optimization is an effective way to learn building-change detectors without manual annotations.

\textbf{Limitations and Future Work.} Despite its strong performance, SST-CD assumes that building changes dominate the stable signals. This may affect variance-based selection in scenes with large-scale natural variations or gradual long-term construction, where unstructured signals can exhibit misleading stability. We leave this issue for future work.

\bibliographystyle{IEEEtran} 
\bibliography{references}

\end{document}